\documentclass[conference,12pt]{IEEEtran}

\IEEEoverridecommandlockouts
\usepackage{cite}
\usepackage{amsmath,amssymb,amsfonts}
\usepackage{algorithmic}
\usepackage{graphicx}
\usepackage{textcomp}
\usepackage{xcolor}
\usepackage{balance}
\begin{document}

\title{TrackNet: A Deep Learning Network for Tracking High-speed and Tiny Objects in Sports Applications}

\author{
{Yu-Chuan Huang}
{ ~ I-No Liao}
{ ~ Ching-Hsuan Chen}
{ ~ Ts\`i-U\'i \.Ik}$^{\ast}$
{ ~ Wen-Chih Peng}
\\
Department of Computer Science, College of Computer Science \\
National Chiao Tung University \\
1001 University Road, Hsinchu City 30010, Taiwan \\
$^{\ast}$Email: cwyi@nctu.edu.tw}

\maketitle

\begin{abstract}
Ball trajectory data are one of the most fundamental and useful information in the evaluation of players' performance and analysis of game strategies. Although vision-based object tracking techniques have been developed to analyze sport competition videos, it is still challenging to recognize and position a high-speed and tiny ball accurately. In this paper, we develop a deep learning network, called TrackNet, to track the tennis ball from broadcast videos in which the ball images are small, blurry, and sometimes with afterimage tracks or even invisible. The proposed heatmap-based deep learning network is trained to not only recognize the ball image from a single frame but also learn flying patterns from consecutive frames. TrackNet takes images with the size of $640\times360$ to generate a detection heatmap from either a single frame or several consecutive frames to position the ball and can achieve high precision even on public domain videos. The network is evaluated on the video of the men's singles final at the 2017 Summer Universiade, which is available on YouTube. The precision, recall, and F1-measure of TrackNet reach $99.7\%$, $97.3\%$, and $98.5\%$, respectively. To prevent overfitting, 9 additional videos are partially labeled together with a subset from the previous dataset to implement 10-fold cross validation, and the precision, recall, and F1-measure are $95.3\%$, $75.7\%$, and $84.3\%$, respectively. A conventional image processing algorithm is also implemented to compare with TrackNet. Our experiments indicate that TrackNet outperforms conventional method by a big margin and achieves exceptional ball tracking performance. The dataset and demo video are available at https://nol.cs.nctu.edu.tw/ndo3je6av9/.
\end{abstract}


\section{Introduction}
\label{sec:Introduction}
Video considered as logs of visual sensors contains a large amount of information. Information extraction from videos has become a hot research topic in the areas of image processing and deep learning. In the applications of sports analyzing and athletes training, videos are helpful in the post-game review and tactical analysis. In professional sports, high-end cameras have been used to record high resolution and high frame rate videos and combined with image processing for referee assistance or data collection. However, this solution requires enormous resources and is not affordable for individuals or amateurs. Developing a low-cost solution for data acquisition from broadcast videos will be significant for massive sports data collection.

Ball trajectory data are one of the most fundamental and useful information for game analysis. However, for some sports such as tennis, badminton, baseball, etc.,  the ball is not only small but also may fly as fast as several hundred kilometers per hour, resulting in tiny and blurry images. That makes the ball tracking task becomes more challenging than other sports. In this paper, we design a heatmap-based deep learning network, called TrackNet, to precisely position ball of tennis and badminton on broadcast videos or videos recorded by consumer's devices such as smartphones. TrackNet overcomes the issues of blurry and remnant images and can even detect occluded ball by learning its trajectory patterns. The proposed network can be applied to other ball-based sports and help both amateurs and professional teams collect data with a moderate budget.

Conventional image recognition is usually based on the object's appearance features such as shape, color, size, etc., or statistical features such as HOG, SIFT, etc. Due to a relatively long shutter time of consumer or prosumer cameras, images of high-speed objects are prone to suffer from afterimage or blur issues, resulting in poor image recognition accuracy. The performance of ball tracking can be improved by pairing candidates from frame to frame according to trajectory models to find the most possible one \cite{Archana2015Object}. In addition, a classical technique in image processing to improve image quality is by fusing multiple low-quality images. Based on the above observations, instead of using the rule-based techniques, we propose to adopt deep learning network to recognize the shape of the ball and learn the trajectory patterns by applying multiple consecutive frames to solve the mentioned issues.

Object classification and detection are two of the earliest studies in deep learning. VGG-16 \cite{Simonyan2014Very} is one of the most popular networks for feature map encoding. To detect and classify multiple objects in an image, the R-CNN family \cite{Girshick2014Rich}\cite{Girshick2015Fast}\cite{Ren2015Faster} structurally examine the picture in two stages. It firstly selects many areas that may contain interesting objects, called Region of Interests (RoIs), and then applies object detection and classification techniques on these regions. However, its performance cannot fulfill the needs of real-time applications. To speed up, the YOLO family \cite{Redmon2016You} develops a one-stage end-to-end approach to detect objects in a limited search space, significantly reducing the computing time. The streamlined version of Tiny YOLO can even run on the Raspberry Pi. Compared to the block-based algorithms, Fully Convolutional Networks (FCN) proceeds pixel-wise classification. To compensate for the size reduction of the feature map during the encoding process, upsampling and DeconvNet \cite{Noh2015Learning} are often used to decode the feature map, generating an original size of the data array.

In this paper, a deep learning network, called TrackNet, is proposed to realize a precise trajectory tracking network. Firstly, VGG-16 is adopted to generate the feature map. Different from other deep learning networks, TrackNet can take multiple consecutive frames as input. In this way, TrackNet learns not only the features of the ball but also the characteristics of ball trajectories to enhance its capability of object recognition and positioning. Since images are downsampled and encoded by pooling layers, the network follows the upsampling mechanism of FCN to generate the heatmap for object detection. At last, the position of our target object is calculated based on the heatmap generated by the deep learning network. To meet the characteristics of tennis and badminton games, our calculation and evaluation are based on the assumption that there is at most one ball on the court.

To evaluate the proposed network, we have labeled $20,844$ frames from the broadcast of men's singles final at the 2017 Summer Universiade. To assess the performance of the proposed consecutive input frames technique, both single-frame and multiple-frame versions of TrackNet are implemented. Along with the conventional image recognition algorithm \cite{Archana2015Object}, a comprehensive comparison among different models is performed. Experiments indicate that the proposed TrackNet outperforms the conventional image recognition algorithm and effectively locates fast-moving tennis ball from broadcast sport competition videos. Moreover, to prevent the notorious overfitting issue that happens frequently in deep learning solutions, additional data from 9 tennis games on different courts are added to the training dataset, including grass court, red clay court, hard court, etc. Additionally, to explore the model extensibility, badminton tracking by TrackNet is evaluated. We have labeled $18,242$ frames from the video of 2018 Indonesia Open Final - TAI Tzu Ying vs CHEN YuFei. Although badminton travels much faster than tennis, our experimental results exhibit a decent performance.

The critical contribution of TrackNet comes from its capability of precisely tracking fast-moving and tiny objects by learning the dynamic behavior of the trajectory. In the tennis tracking application, 10-fold cross validation results in an outstanding performance of $95.3\%$ precision, $75.7\%$ recall, and $84.3\%$ F1-measure. Such capability shows great potential in expanding the variety of computer vision applications. The rest of the paper is organized as follows. Section \ref{sec:RelatedWorks} provides an introduction to the relevant researches and the convolutional neural network. Section \ref{sec:Dataset} introduces the datasets used in this paper. Section \ref{sec:TrackNet} elaborates the proposed deep learning network and Gaussian heatmap techniques. Section \ref{sec:Experiments} provides experimental results and performance evaluation. At last, Section \ref{sec:Conclusion} concludes this paper.

\section{Related Works}
\label{sec:RelatedWorks}
In recent years, the analysis of player performance and game tactics based on the trajectory data of balls and players has received more and more attention \cite{Chen2012Ball}\cite{Wang2014Take}\cite{Fu2011Screen-strategy}\cite{Myint2015Tracking}. Many tracking algorithms and systems have been developed to compute and collect the trajectory data. Current commercial solutions mainly rely on high resolution and high frame rate video, resulting in high hardware investment. For example, the Hawk-Eye system \cite{HawkEye} has been extensively used in professional competitions to calculate ball trajectories and assist the referee in clarifying controversial calls through 3D visual depictions. Nonetheless, the system has to deploy high-end cameras with dedicated operators at selected locations and angles. The expense is too high for non-professional teams.

Attempting to position the ball from sports competition videos has been studied for years. However, since the ball size is relatively small, it is prone to be confused with objects having similar color or shape, causing false positives. Furthermore, due to the high moving speed of the ball, the resulting image is usually blurry, inducing false negatives. By exploring the trajectory pattern from consecutive frames, the ball positioning can be effectively improved. In addition, the flight trajectory itself possesses important information and is a subject in many pieces of research \cite{yu2004trajectory}. For instance, combining multiple cameras with 3D technology for tennis detection \cite{Reno2016Real}, tracking tennis by particle filter in low-quality films \cite{Yan2005Tennis}, and adopting two-layer data association approach to calculate the most likely ball trajectory from the results of failure detection in the frame-by-frame image processing \cite{Zhou2015Tennis} are enlightening studies.

The success of deep learning techniques in image classification \cite{Simonyan2014Very}\cite{Krizhevsky2012Imagenet} encourages more researchers to adopt these methods to solve various problems such as object detection and interception \cite{Ren2015Faster}\cite{Redmon2016You}\cite{Ronneberger2015U}, computer games, network security, activity recognition \cite{Jiang2015Human}\cite{Chen2015Deep}, text and image semantic analysis, and smart stores. The infrastructure of the deep learning network is a structured and huge convolutional neural network trained with a large amount of labeled data. The most common operations of CNNs include convolution, rectifier, pooling/down-sampling, and deconvolution/up-sampling. A softmax layer is usually used as the output layer. For example, the widely used VGG-16 \cite{Simonyan2014Very} mainly consists of convolutional, maximum pooling, and ReLU layers. Conceptually, front-end layers learn to identify simple geometric features, and back-end layers are trained to identify object features.

\begin{figure}[pth]
\centering
\includegraphics[angle=0, width=1.6in,keepaspectratio,clip]{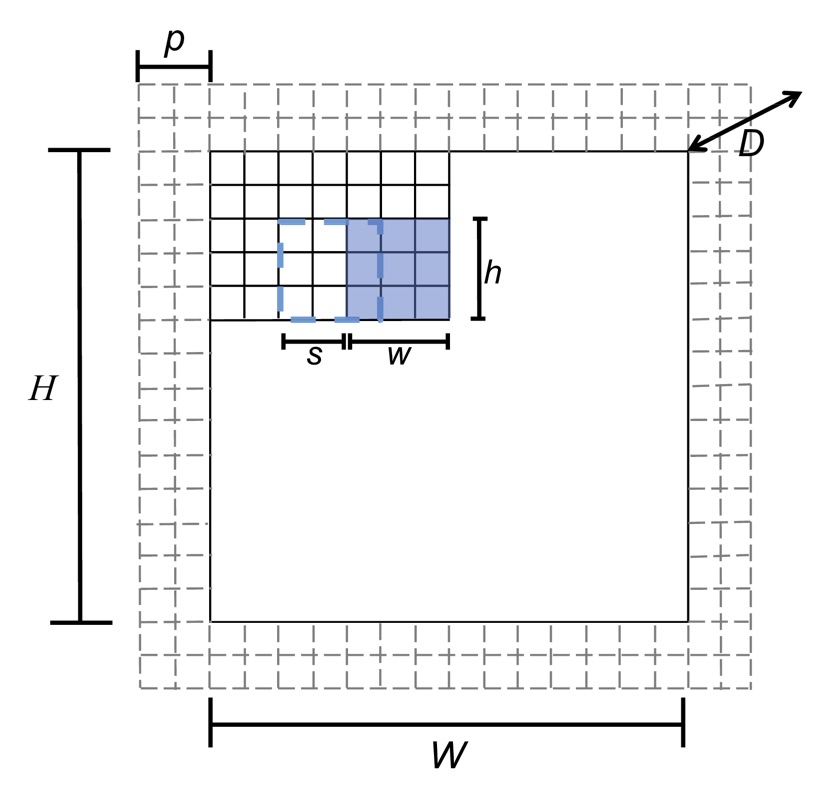}
\hfill\caption{Convolution operation in deep learning networks.}
\label{f_Convolution}
\end{figure}

In CNNs, each layer is a $W\times H\times D$ data array. $W$, $H$, and $D$ denote the width, height, and depth of the data array, respectively. The convolution operation is a filter with a kernel of size $w\times h\times D$ across the $W\times H$ range with the stride parameter $s$ being set as $1$ in many applications. To avoid information loss near the boundary or maintain the size of the output data array, columns and rows of the data array can be padded with zero by setting the padding parameter $p$. Figure \ref{f_Convolution} depicts the relevant parameters of the convolution operation. Let $W'$ and $H'$ denote the width and height of the next layer. Then, 
\[
W^{\prime}=\frac{W+2p-w}{s}+1\text{ and }H^{\prime}=\frac{H+2p-h}{s}+1\text{.}
\]

Since the convolution operation is linear and cannot effectively capture nonlinear behaviors, an activation function called rectifier is introduced to capture nonlinear behaviors. The Rectified Linear Unit (ReLU) is the most commonly used activation function in deep learning models. If the input value is negative, the function returns $0$; otherwise, the function returns the input value. ReLU can be expressed as $f\left(x\right)=\max(0,x)$. Maximum pooling provides the functionality of down-sampling and feature fusion. Maximum pooling fuses features by encoding data via down-sampling. The block of data will be represented only by the largest one. After pooling, the data size is reduced. On the other hand, to achieve pixel-by-pixel classification, up-sampling is necessary to reconstruct an output with the same size as the original image \cite{Badrinarayanan2015Segnet}\cite{Long2015Fully}. In up-sampling, samples are duplicated to expand the data size. Batch normalization is a widely used technique to speed up the training process. Each $W\times H$ data array is independently standardized into a normal distribution.

Backward propagation is commonly used in training neural networks to learn the filter coefficients. Firstly, forward propagation is performed to have a preliminary prediction. Then, compared the prediction with the ground truth, a loss function will be evaluated. Finally, the weights of the model, i.e., the filter coefficients, are updated according to the loss by the gradient descent method. Chain rule is adopted to calculate the gradient of the loss function layer by layer. The process will be repeated again and again until a certain number of repetitions is reached or the loss falls below an acceptable threshold. The design of the loss function is an important factor that affects the training efficiency and the performance of the network. Commonly used loss functions include Root Mean Square Error (RMSE) and cross-entropy.

In this paper, we propose a deep learning network named TrackNet to detect tennis and badminton on broadcast sport competition videos. By training with consecutive input frames, TrackNet can not only recognize the ball but also learn its trajectory pattern. A heatmap which is ideally a Gaussian distribution centered on the ball image is then generated by TrackNet to indicate the position of the ball. The idea of exploiting heatmap for object detection has been adopted in many studies \cite{Belagiannis2017Recurrent}\cite{Pfister2015Flowing}. 

To compare and evaluate the performance of TrackNet, we implement Archana’s algorithm \cite{Archana2015Object} which uses conventional image processing techniques to detect tennis ball. Archana's algorithm firstly smooths the image of each frame by a median filter to remove noise. After a background model is calculated, background subtraction is performed to obtain the foreground. Then, the difference between frames by logical AND operation is examined to identify fast-moving foreground objects. Those objects are compared with shape, size, and aspect ratio of the tennis ball and selected by applying dilation and erosion to generate candidates. To filter out wrong candidates, in our implementation, a fully-connected neural network is trained to classify candidates into positive and negative categories. The one that has the highest probability in the positive category is selected, indicating the position of the ball.

\section{Dataset}
\label{sec:Dataset}
Our first dataset is from the broadcast video of the tennis men's singles final at the 2017 Summer Universiade. The resolution, frame rate, and video length are $1280\times720$, 30 fps, and 75 minutes, respectively. By screening out unrelated frames, 81 game-related clips are segmented and each of them records a complete play, starting from ball serving to score. There are $20,844$ frames in total. Each frame possesses the following attributes: "Frame Name", "Visibility Class", "X", "Y", and "Trajectory Pattern". Table \ref{t_LabelFile} is pieces of label files.

\begin{table}[ptb]
\caption{Segments of label files. }
\label{t_LabelFile}
\begin{center}
$\vdots$
\par
0008.jpg, 2, 727, 447, 0
\par
0009.jpg, 1, 735, 457, 0
\par
0010.jpg, 1, 722, 433, 1
\par
0011.jpg, 1, 707, 403, 0
\par
$\vdots$
\par
0029.jpg, 1, 555, 220, 0
\par
0030.jpg, 1, 550, 218, 2
\par
0031.jpg, 1, 547, 206, 0
\par
$\vdots$
\end{center}
\end{table}

\begin{figure}[pth]
\centering
\includegraphics[angle=0, width=3.3in,keepaspectratio,clip] {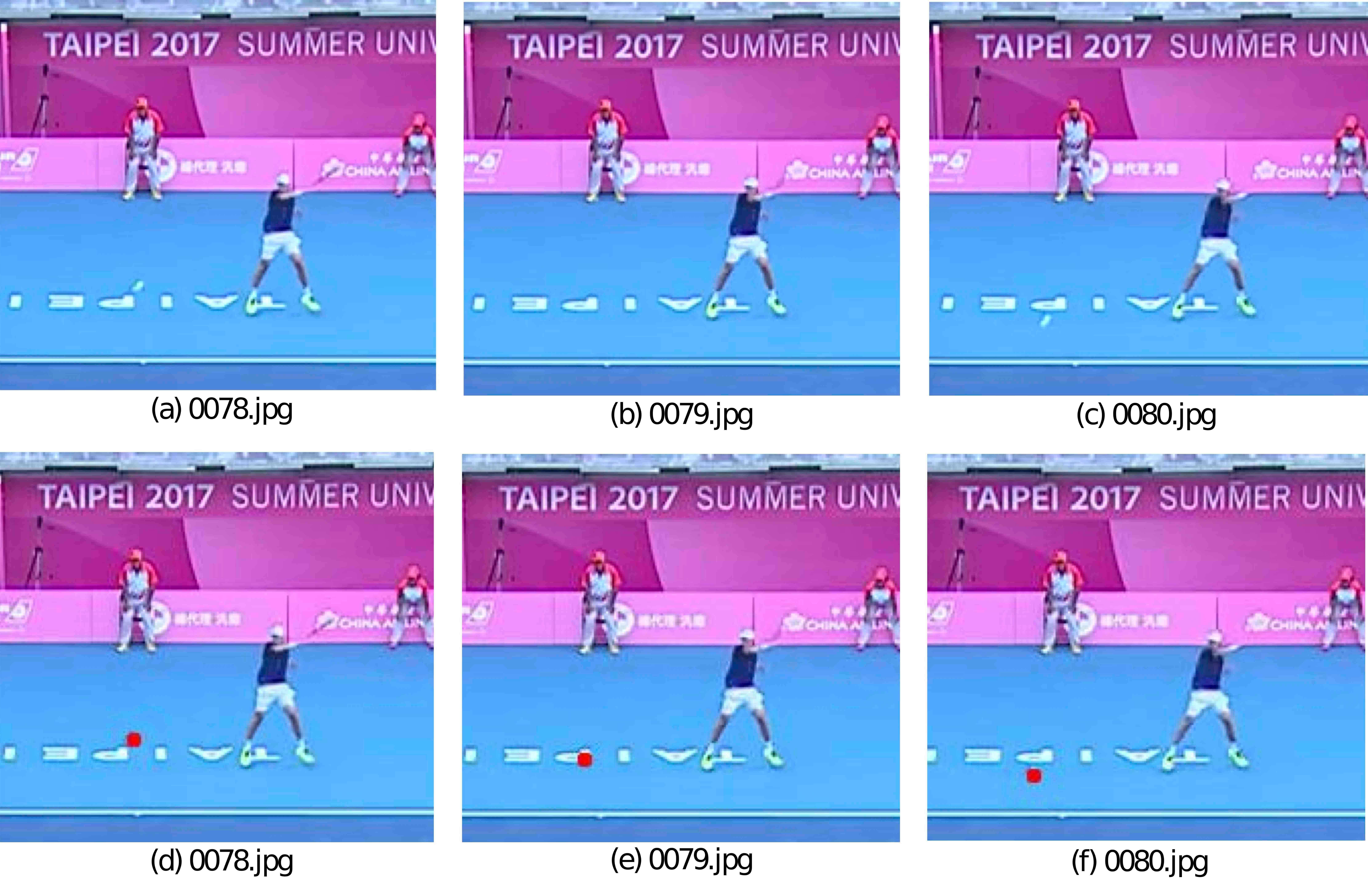}
\hfill\caption{The ball image is hardly visible.}
\label{f_InvisibleBall}
\end{figure}

\begin{figure}[pth]
\centering
\includegraphics[angle=0, width=3.3in,keepaspectratio,clip]{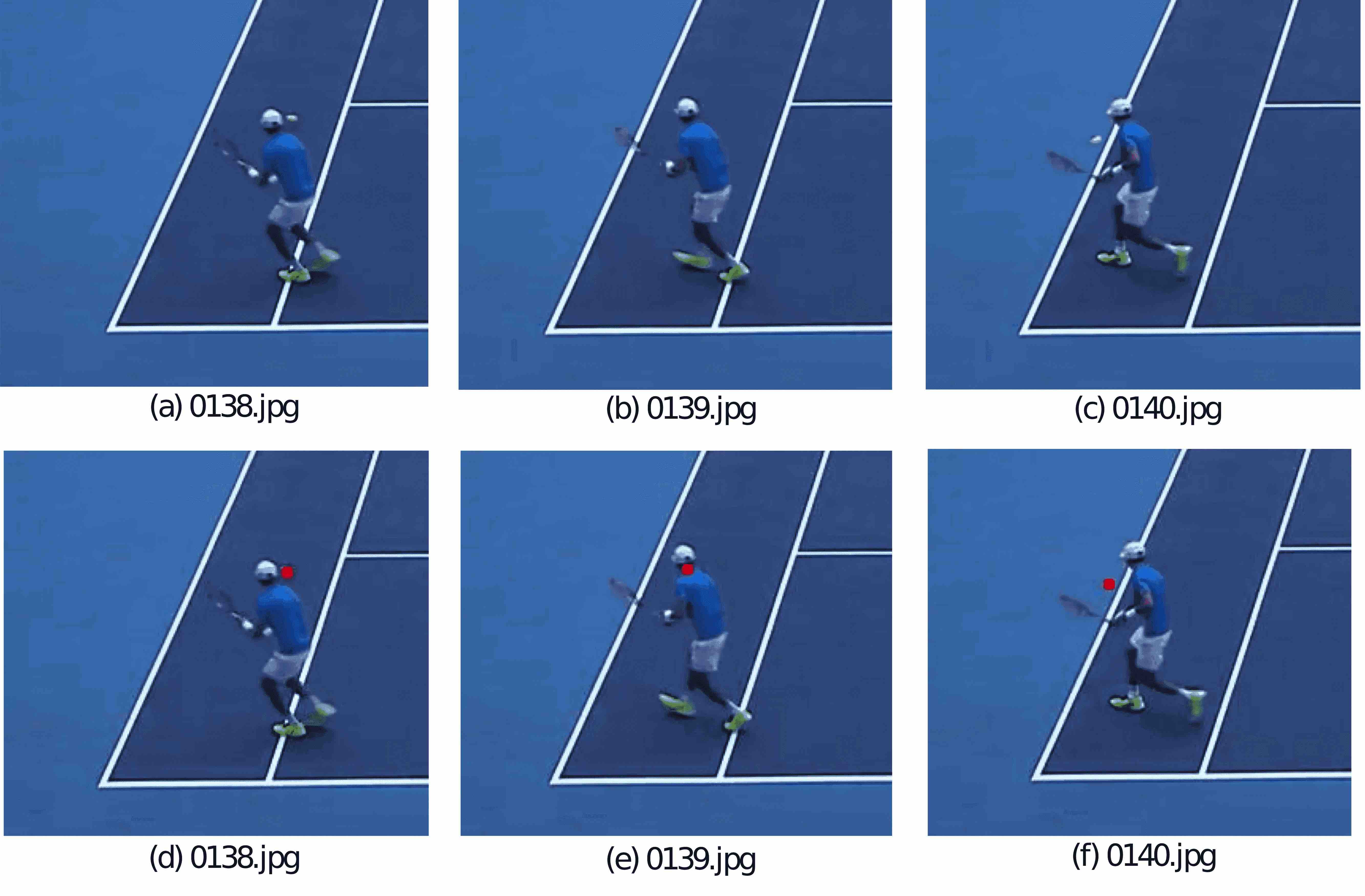}
\hfill\caption{The ball is occluded by the player.}
\label{f_OccludedBall}
\end{figure}

"Frame Name" is the name of the frame files. "Visibility Class", VC for short, indicates the visibility of the ball in each frame. The possible values are 0, 1, 2, and 3. $VC=0$ implies the ball is not within the frame. $VC=1$ implies the ball can be easily identified. $VC=2$ implies the ball is in the frame but can not be easily identified. For example, as shown in Figure \ref{f_InvisibleBall}, the ball in 0079.jpg is hardly visible since the color of the tennis ball is similar to the text "Taipei" on the court. However, with the help of neighboring frames, 0078.jpg and 0080.jpg, the unclear ball position of 0079.jpg can be labeled. Figure \ref{f_InvisibleBall} (d), (e), and (f) illustrate the labeling results. $VC=3$ implies the ball is occluded by other objects. For example, as shown in Figure \ref{f_OccludedBall}, the ball in 0139.jpg is occluded by the player. Similarly, based on the information from neighboring frames, 0138.jpg and 0140.jpg, the ball position of 0139.jpg can be estimated. Figure \ref{f_OccludedBall} (d), (e), and (f) illustrate the labeling results. In the dataset, the number of frames of $VC=0, 1, 2, 3$ are 659, 18035, 2143, and 7, respectively.

\begin{figure}[pth]
\centering
\includegraphics[angle=0, width=3.3in,keepaspectratio,clip] {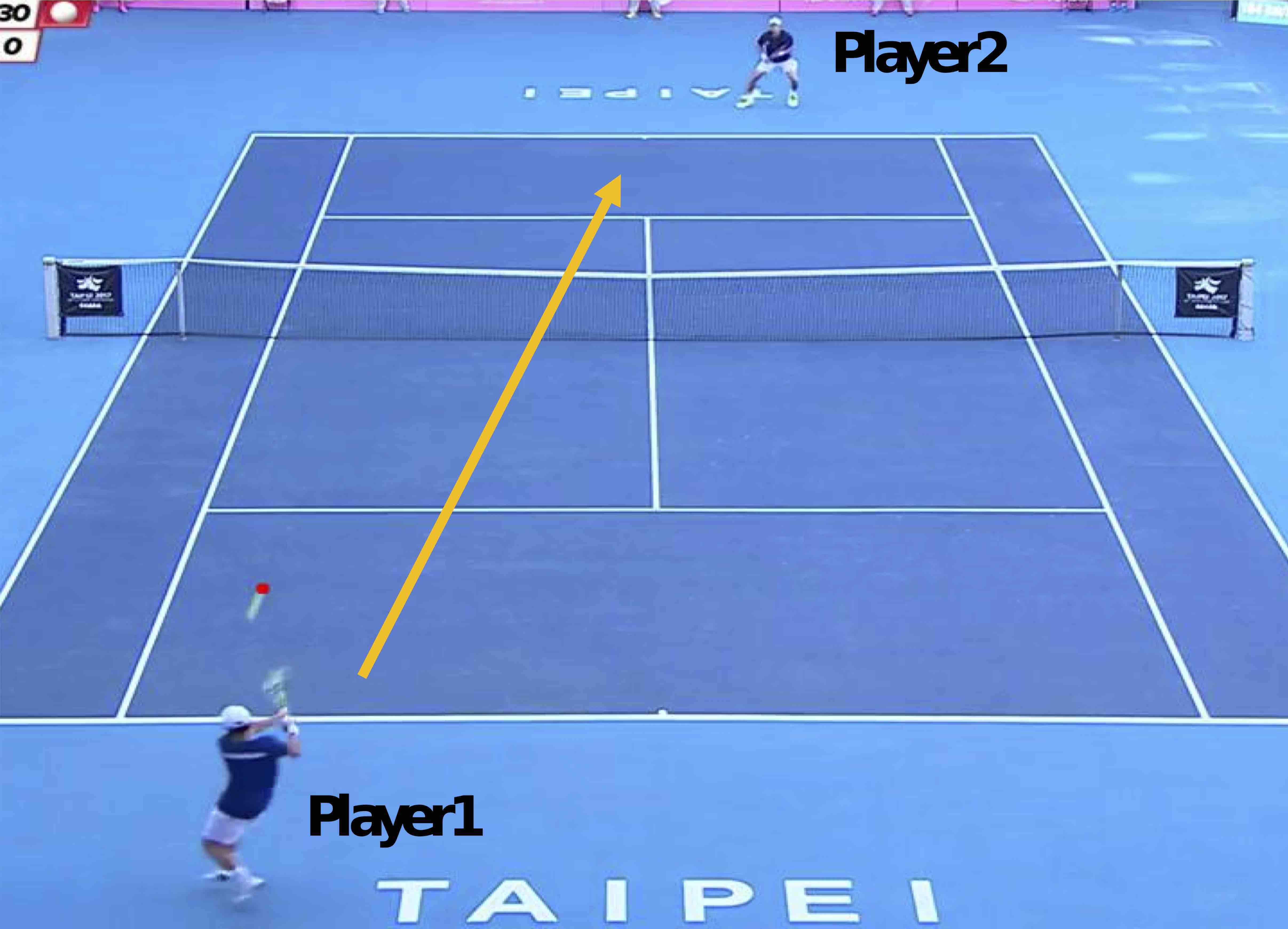}
\hfill\caption{An example of the prolonged tennis trace.}
\label{f_ProlongedTrack}
\end{figure}

"X" and "Y" indicate the coordinate of tennis in the pixel coordinate. Due to the high moving speed, tennis images in the broadcast video may be blurry and even have afterimage trace. In such cases, "X" and "Y" are considered as the latest position of the ball's trace. For example, as shown in Figure \ref{f_ProlongedTrack}, the ball is flying from Player1 to Player2 with a prolonged trace and the red dot indicates the labeled coordinate.

\begin{figure}[pth]
\centering
\includegraphics[angle=0, width=3.3in,keepaspectratio,clip] {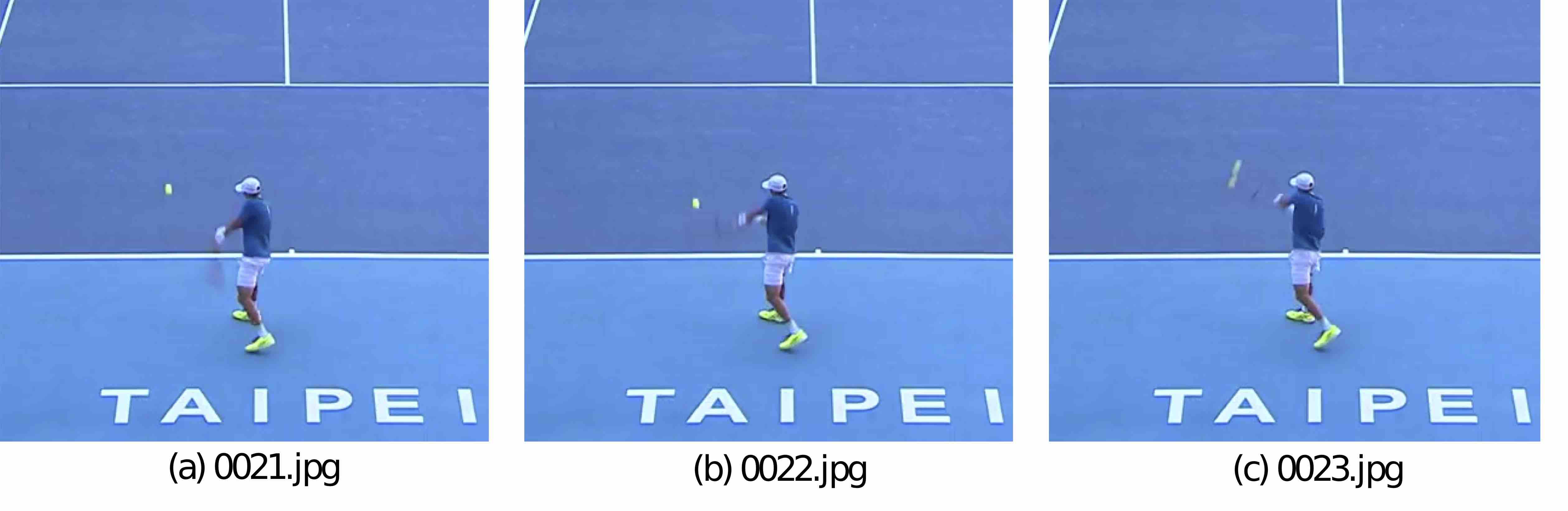}
\hfill\caption{A hit case: (a) and (b) are labeled as flying, and (c) is labeled as hit.}
\label{f_Hit}
\end{figure}

\begin{figure}[pth]
\centering
\includegraphics[angle=0, width=3.3in,keepaspectratio,clip] {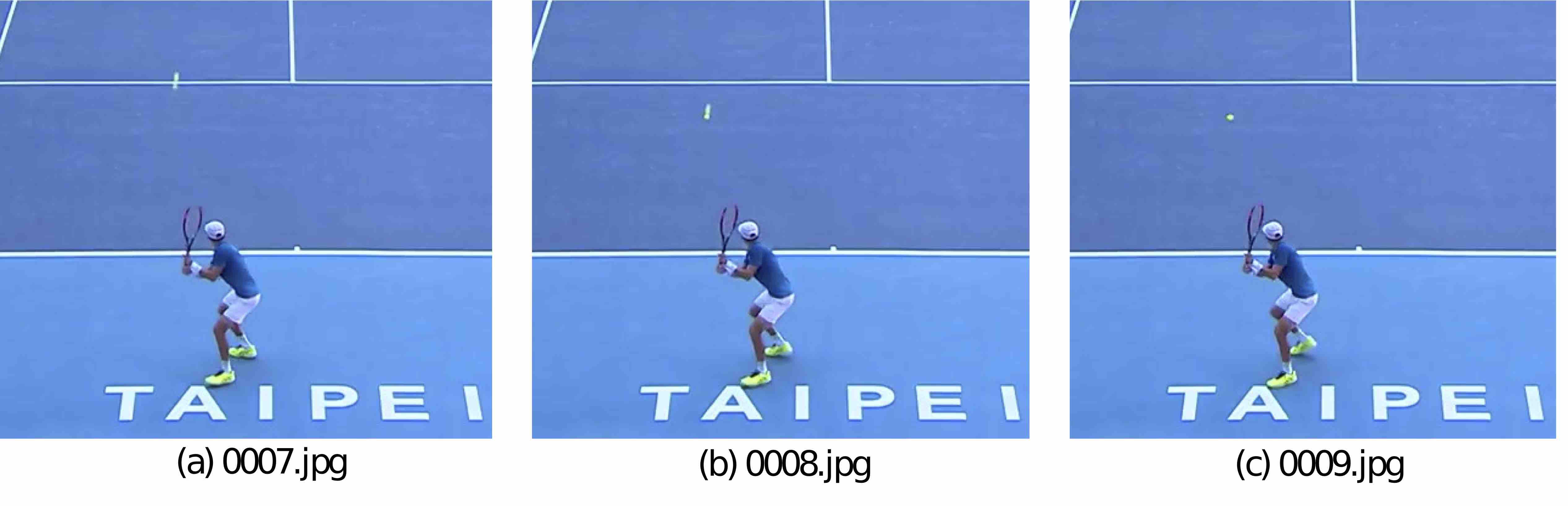}
\hfill\caption{A bouncing case: (a) and (b) are labeled as flying, and (c) is labeled as bouncing.}
\label{f_Bouncing}
\end{figure}

"Trajectory Pattern" indicates the ball movement types and are classified into three categories: flying, hit, and bouncing. They are labeled by 0, 1, and 2, respectively. Figure \ref{f_Hit} is an example of striking a ball. The ball is flying at 0021.jpg and 0022.jpg. At 0023.jpg, the ball is labeled as hit. Figure \ref{f_Bouncing} shows a bouncing case. The ball has not reached the ground at 0007.jpg and 0008.jpg. At 0009.jpg, the ball hits the ground and is labeled as bouncing.

To enrich the variety of training dataset, additional $16,118$ frames are collected. These frames came from $9$ videos recorded at different tennis courts, including grass court, red clay court, hard court etc. By learning diverse scenarios, the deep learning model is expected to recognize tennis ball at various courts. That increases the robustness of the model. Further details will be presented in Section \ref{sec:Experiments}.

In addition to tennis, to explore the versatility of the proposed TrackNet in the applications of high-speed and tiny objects tracking, a trial run on badminton match video is performed. Tracking badminton is more challenging than tracking tennis since the speed of badminton is much faster than tennis. The fastest serve according to the official records from the Association of Tennis Professionals is John Isner's $253$ kilometers per hour at the 2016 Davis Cup. On the other hand, the fastest badminton hit in competition is Lee Chong Wei's $417$ kilometers per hour smash at the 2017 Japan Open according to Guinness World Records, which is over $1.6$ times faster than tennis. Besides, in professional competitions, the speed of badminton is frequently over $300$ kilometers per hour. The faster the object moves, the more difficult it is to be tracked. Hence, it is expected that the performance will degrade for badminton compared with tennis.

Our badminton dataset comes from a video of the badminton competition of 2018 Indonesia Open Final - TAI Tzu Ying vs CHEN YuFei. The resolution is $1280\times720$ and the frame rate is 30 fps. Similarly, unrelated frames such as commercial or highlight replays are screened out. The resulting total number of frames is $18,242$. We label each frame with the following attributes: "Frame Name", "Visibility Class", "X", and "Y".

In badminton dataset, "Visibility Class" is classified into two categories, $VC=0$ and $VC=1$. $VC=0$ means the ball is not in the frame and $VC=1$ means the ball is in the frame. Unlike our tennis dataset, we do not classify $VC=2$ and $VC=3$ categories since the badminton moves so fast that blurry image happens very frequently. Therefore, in the badminton dataset, $VC=1$ includes all status of badminton as long as the ball is within the frame no matter it is clearly visible or hardly visible.

\begin{figure}[pth]
\centering
\includegraphics[angle=0, width=3.3in,keepaspectratio,clip] {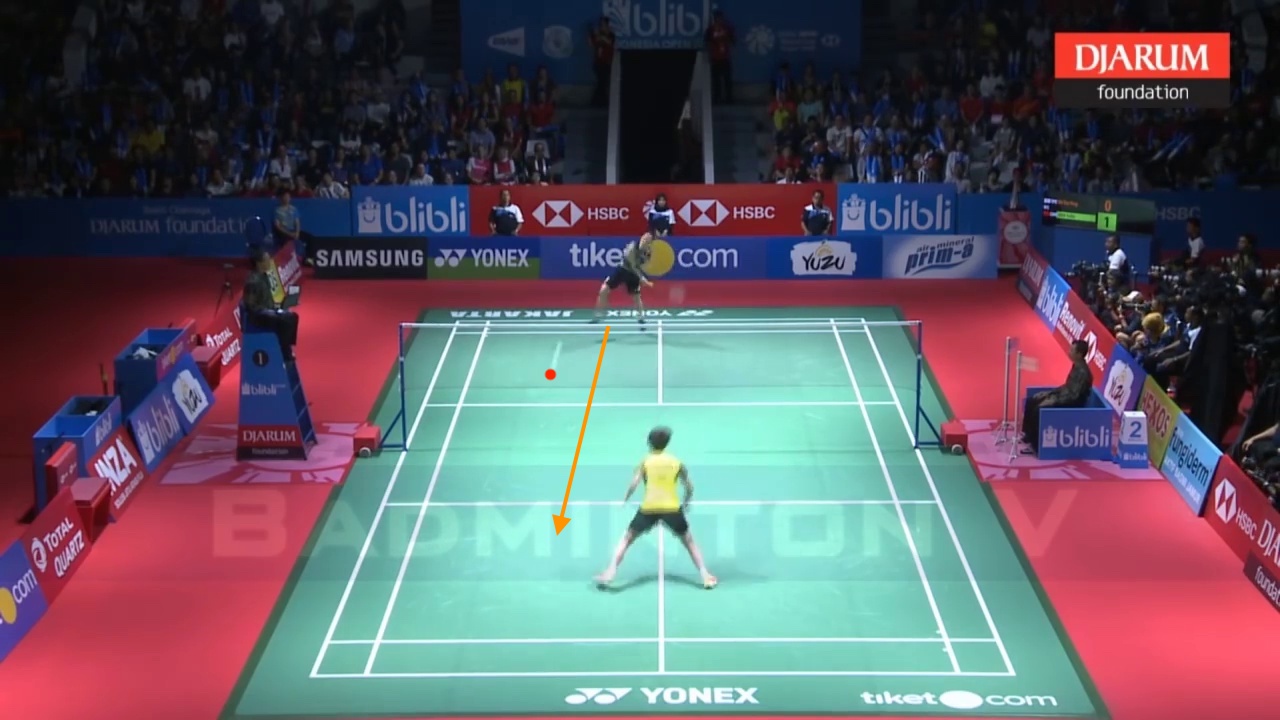}
\hfill\caption{An example of the prolonged badminton trace.}
\label{f_ProlongedTrackBadminton}
\end{figure}

"X" and "Y" indicate the coordinate of badminton. Similar to tennis, "X" and "Y" are defined by the latest position of the ball's trace considering its moving direction if the image is prolonged. In badminton video, prolonged trace often happens and sometimes we could hardly identify the position of the ball. An example of how we label the prolonged images is shown in Figure \ref{f_ProlongedTrackBadminton}.

\section{TrackNet}
\label{sec:TrackNet}

\begin{figure}[pth]
\centering
\includegraphics[angle=0, width=2in,keepaspectratio,clip]{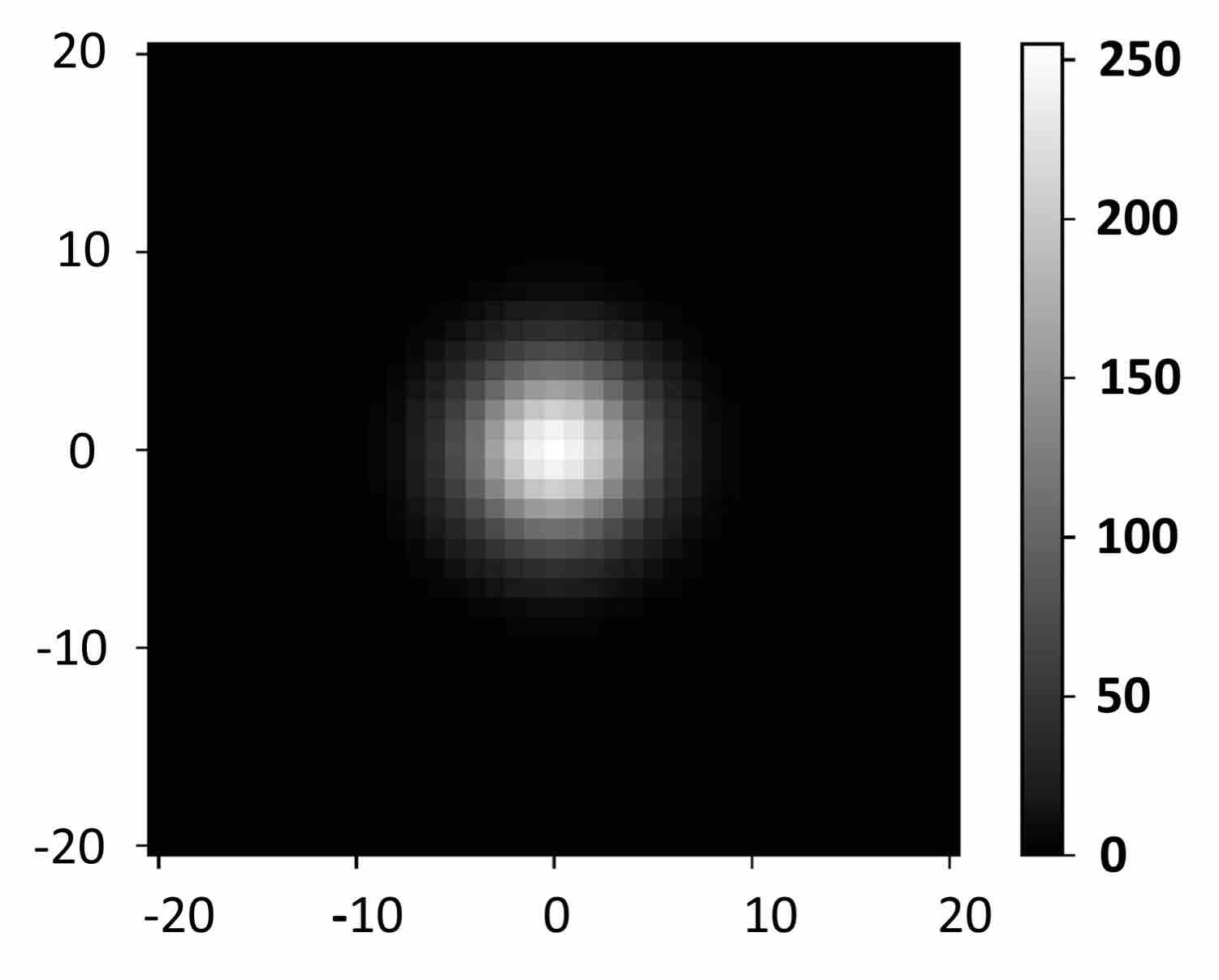}
\hfill\caption{An example of the detection heatmap.}
\label{f_Heatmap}
\end{figure}

TrackNet is composed of a convolutional neural network (CNN) followed by a deconvolutional neural network (DeconvNet) \cite{Noh2015Learning}. It takes consecutive frames to generate a heatmap indicating the position of the object. The number of input frames is a network parameter. One input frame is considered the conventional CNN network. TrackNet with more than one input frame can improve the moving object detection by learning the trajectory pattern. For the purpose of evaluation, two networks are implemented. One is with single frame input, and the other is with three consecutive frames input.

\begin{figure*}[pth]
\centering
\includegraphics[angle=0, width=5in,keepaspectratio,clip]{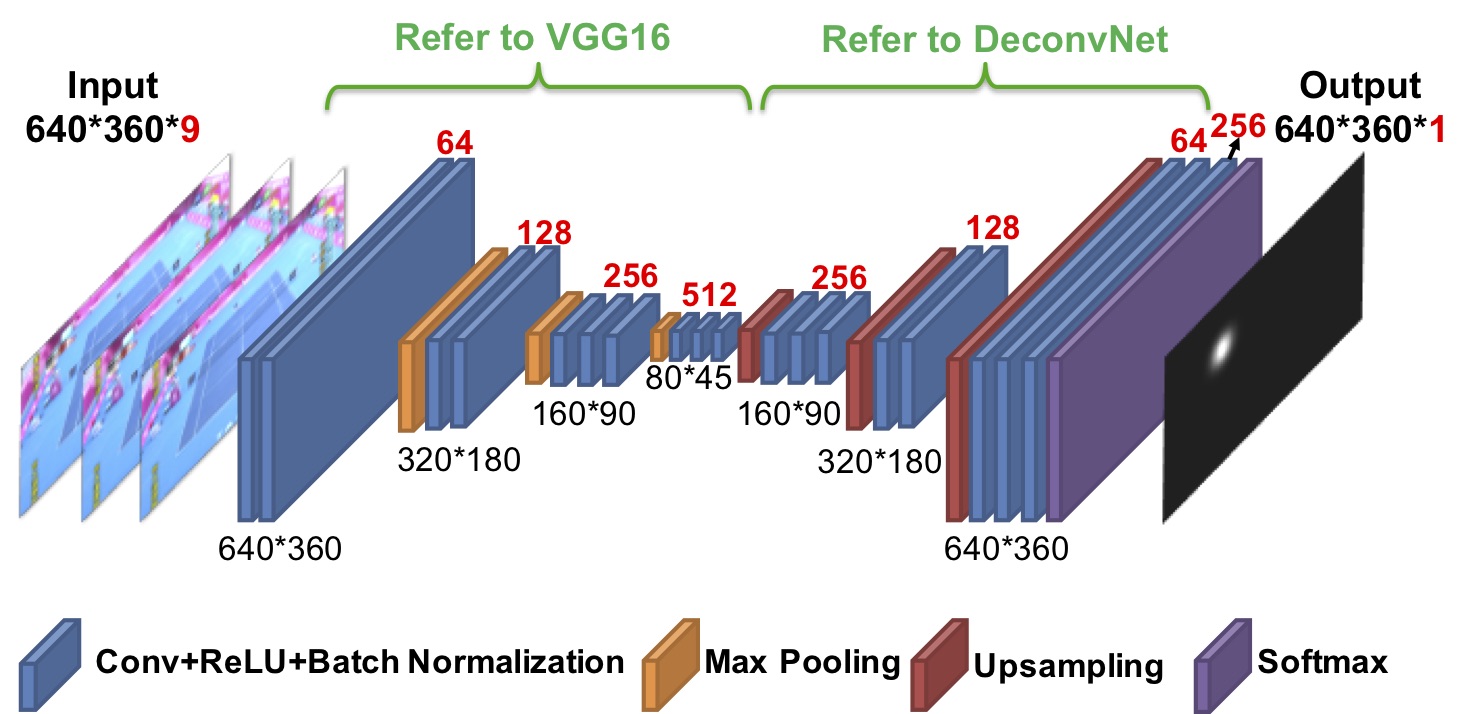}
\hfill\caption{The architecture of the proposed TrackNet.}
\label{f_TrackNet}
\end{figure*}

TrackNet utilizes the heatmap-based CNN which has been proved useful in several applications \cite{Belagiannis2017Recurrent}\cite{Pfister2015Flowing}. TrackNet is trained to generate a probability-like detection heatmap having the same resolution as the input frames. The ground truth of the heatmap is an amplified 2D Gaussian distribution located at the center of the tennis ball. The coordinates of the ball are available in the labeled dataset and the variance of the Gaussian distribution refers to the diameter of tennis ball images. Let $\left(x_{0} ,y_{0}\right)$ be the ball center and the heatmap function is expressed as
\[
G\left(x,y\right)=\left\lfloor \left(  \frac{1}{2\pi\sigma^{2}}e^{-\frac{\left(x-x_{0}\right)^{2}+\left(y-y_{0}\right)^{2}} {2\sigma^{2}}}\right)\left(2\pi\sigma^{2}\cdot255\right)\right\rfloor,
\]
where the first part is a Gaussian distribution centered at $\left(x_{0},y_{0}\right)$ with variance of $\sigma^{2}$, and the second part scales the value to the range of $\left[0, 255\right]$. $\sigma^{2}=10$ is used in our implementation since the average ball radius is about 5 pixels, roughly corresponding to the region of $G\left(x,y\right)\geq128$. Figure \ref{f_Heatmap} is a visualized heatmap function of a tennis ball.

The implementation details of TrackNet is illustrated in Figure \ref{f_TrackNet} and Table \ref{t_TrackNet}. The input of the proposed network can be some number of consecutive frames. The first 13 layers refer to the design of the first 13 layers of VGG-16 \cite{Simonyan2014Very} for object classification. The 14-24 layers refer to DeconvNet \cite{Noh2015Learning} for semantic segmentation. To realize the pixel-wise prediction, upsampling is applied to recover the information loss from maximum pooling layers. Symmetric numbers of upsampling layers and maximum pooling layers are implemented.

\begin{table}[ptb]
\caption{Network parameters of TrackNet.}
\label{t_TrackNet}
\begin{center}
\begin{tabular}
[c]{|c|c|c|c|c|c|}\hline
Layer & Filter Size & Depth & Padding & Stride & Activation\\\hline
Conv1 & $3\times3$ & 64 & 2 & 1 & ReLU+BN\\\hline
Conv2 & $3\times3$ & 64 & 2 & 1 & ReLU+BN\\\hline
Pool1 & \multicolumn{5}{|c|}{$2\times2$ max pooling and $Stride=2$}\\\hline
Conv3 & $3\times3$ & 128 & 2 & 1 & ReLU+BN\\\hline
Conv4 & $3\times3$ & 128 & 2 & 1 & ReLU+BN\\\hline
Pool2 & \multicolumn{5}{|c|}{$2\times2$ max pooling and $Stride=2$}\\\hline
Conv5 & $3\times3$ & 256 & 2 & 1 & ReLU+BN\\\hline
Conv6 & $3\times3$ & 256 & 2 & 1 & ReLU+BN\\\hline
Conv7 & $3\times3$ & 256 & 2 & 1 & ReLU+BN\\\hline
Pool3 & \multicolumn{5}{|c|}{$2\times2$ max pooling and $Stride=2$}\\\hline
Conv8 & $3\times3$ & 512 & 2 & 1 & ReLU+BN\\\hline
Conv9 & $3\times3$ & 512 & 2 & 1 & ReLU+BN\\\hline
Conv10 & $3\times3$ & 512 & 2 & 1 & ReLU+BN\\\hline
UpS1 & \multicolumn{5}{|c|}{$2\times2$ upsampling}\\\hline
Conv11 & $3\times3$ & 512 & 2 & 1 & ReLU+BN\\\hline
Conv12 & $3\times3$ & 512 & 2 & 1 & ReLU+BN\\\hline
Conv13 & $3\times3$ & 512 & 2 & 1 & ReLU+BN\\\hline
UpS2 & \multicolumn{5}{|c|}{$2\times2$ upsampling}\\\hline
Conv14 & $3\times3$ & 128 & 2 & 1 & ReLU+BN\\\hline
Conv15 & $3\times3$ & 128 & 2 & 1 & ReLU+BN\\\hline
UpS3 & \multicolumn{5}{|c|}{$2\times2$ upsampling}\\\hline
Conv16 & $3\times3$ & 64 & 2 & 1 & ReLU+BN\\\hline
Conv17 & $3\times3$ & 64 & 2 & 1 & ReLU+BN\\\hline
Conv18 & $3\times3$ & 256 & 2 & 1 & ReLU+BN\\\hline
\multicolumn{6}{|c|}{Softmax}\\\hline
\end{tabular}
\end{center}
\end{table}

The final black-white binary detection heatmap is not directly available at the output of the deep learning network. The network outputs a detection heatmap that has continuous values within the range of $\left[0, 255\right]$ for each pixel. Let $L\left(i,j,k\right)$ denote the data array of coordinates within $(0,0)\leq(i,j)\leq(639,359)$ and depth within $0\leq k\leq255$. The softmax layer calculates the probability distribution of depth $k$ from possible 256 grayscale values. Let $P\left(i,j,k\right)$ denote the probability of depth $k$ at $\left(i,j\right)$. The softmax function is given by
\[
P\left(i,j,k\right)  =\frac{e^{L\left(i,j,k\right)}}{\sum_{l=0}^{255}e^{L\left(i,j,l\right)}}.
\]
Based on the probability given by the softmax layer on each pixel, the depth $k$ with the highest probability is selected as the heatmap value of the pixel. For each pixel, let
\[
h\left(i,j\right)  =\arg\max_{k}P\left(i,j,k\right)
\]
denote the softmax layer output at $\left(i,j\right)$, indicating the selected grayscale value at $\left(i,j\right)$. Once the complete continuous detection heatmap is generated, the coordinate of the ball can be determined by the following two steps. The first step is to pixel-wisely convert the heatmap into a black-white binary heatmap by the threshold $t$. If a pixel has a value larger than or equal to $t$, the pixel is set to $255$. On the contrary, if a pixel has a value smaller than $t$, the pixel is set to $0$. Based on the previous discussion regarding the mean radius of a tennis ball, threshold $t$ is set as $128$. The second step is to exploit the Hough Gradient Method \cite{HoughGradientMethod} to find the circle on the black-white binary detection heatmap. If exactly one circle is identified, the centroid of the circle is returned. In other cases, the heatmap is considered no ball detected.

During the training phase, the cross-entropy function is used to calculate the loss function based on $P\left(i,j,k\right)$. The corresponding ground truth function denoted by $Q\left(i,j,k\right)$ is given by
\[
Q\left(  i,j,k\right)  =\left\{
\begin{array}
[c]{l}
1\text{, if }G\left(  i,j\right)=k\text{;}\\
0\text{, otherwise.}
\end{array}
\right.
\]
Let $H_{Q}\left(P\right)$ denote the loss function. Then,
\[
H_{Q}\left(P\right)  =-\sum_{i,j,k}Q\left(i,j,k\right)\log P\left(i,j,k\right).
\]

\section{Experiments}
\label{sec:Experiments}
\begin{figure}[pth]
\centering
\includegraphics[angle=0, width=3in,keepaspectratio,clip]{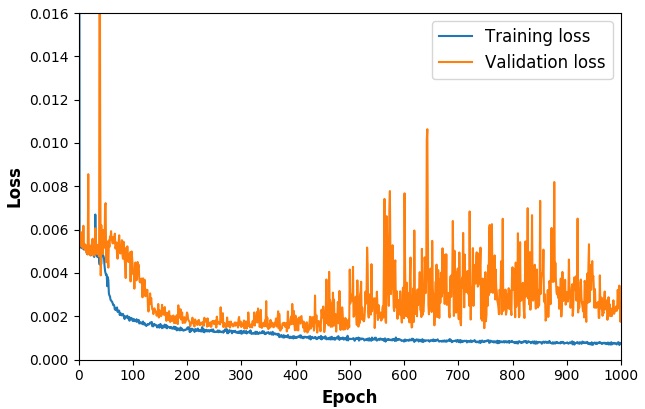}
\hfill\caption{The loss curve of TrackNet model training.}
\label{f_LossPerEpoch}
\end{figure}

The experiment setup is as followed. The tennis dataset elaborated in Section \ref{sec:Dataset} is used to evaluate the performance of Archana's algorithm, a conventional image processing technique, and the proposed TrackNet. The dataset contains $20,844$ frames and is randomly divided to the training set and test set. $70\%$ frames are the training set and $30\%$ frames are the test set. To speed up the training speed, all frames are resized from $1280\times720$ to $640\times360$. To optimize weights of the network, the Adadelta optimizer \cite{Zeiler2012Adadelta} is applied. Table \ref{t_TrainingParameters} summarizes other key parameters. Among these parameters, the number of epochs is one of the most critical factors in model training. Underfitting happens if it is too small, while overfitting happens if it is too large. For TrackNet, the characteristic of loss versus the number of epochs is shown in Figure \ref{f_LossPerEpoch}. Based on the simulation, we select 500 epochs as our optimal value to prevent both underfitting and overfitting.

\begin{table}[ptb]
\caption{Key parameters used in model training.}
\label{t_TrainingParameters}
\begin{center}
\begin{tabular}
[c]{|c|c|}\hline
Parameters & Setting\\\hline
Learning rate & 1.0\\\hline
Batch size & 2\\\hline
Steps per epoch & 200\\\hline
epochs & 500\\\hline
Initial weights & random uniform\\\hline
Range of initial weights & $\left[  -0.05,0.05\right]  $\\\hline
\end{tabular}
\end{center}
\end{table}

\begin{table*}[ptb]
\caption{Performance Summary.}
\label{t_PerformanceSummary}
\begin{center}
\begin{tabular}
[c]{|c|c|c|c|c|c|c|c|c|c|c|c|c|c|c|c|c|}\hline
& \multicolumn{4}{|c|}{Archana's} & \multicolumn{4}{|c|}{TrackNet Model I} & \multicolumn{4}{|c|}{TrackNet Model II} & \multicolumn{4}{|c|}{TrackNet Model II'}\\\hline
& VC0 & VC1 & VC2 & VC3 & VC0 & VC1 & VC2 & VC3 & VC0 & VC1 & VC2 & VC3 & VC0 & VC1 & VC2 & VC3\\\hline
TP & - & 4046 & 418 & 0 & - & 4933 & 497 & 0 & - & 5223 & 565 & 2 & - & 5234 & 598 & 1\\\hline
FP & 201 & 334 & 29 & 1 & 1 & 221 & 20 & 0 & 0 & 3 & 3 & 0 & 4 & 6 & 7 & 2\\\hline
TN & 9 & - & - & - & 195 & - & - & - & 210 & - & - & - & 206 & - & - & -\\\hline
FN & - & 947 & 214 & 6 & - & 241 & 139 & 7 & - & 101 & 93 & 5 & - & 87 & 56 & 4\\\hline
Total & 210 & 5327 & 661 & 7 & 196 & 5395 & 656 & 7 & 210 & 5327 & 661 & 7 & 210 & 5327 & 661 & 7\\\hline
\end{tabular}
\end{center}
\end{table*}

\begin{table}[ptb]
\caption{Accuracy metrics of different models.}
\label{t_Comparsion}
\begin{center}
\begin{tabular}
[c]{|l|c|c|c|}\hline
Model & Precision & Recall & F1-measure\\\hline
Archana's \cite{Archana2015Object} & 92.5\% & 74.5\% & 82.5\%\\\hline
TrackNet Model I & 95.7\% & 89.6\% & 92.5\%\\\hline
TrackNet Model II & 99.8\% & 96.6\% & 98.2\%\\\hline
TrackNet Model II' & 99.7\% & 97.3\% & 98.5\%\\\hline
\end{tabular}
\end{center}
\end{table}

To compare the performance of TrackNet frameworks with one input frame and three consecutive input frames, two versions of TrackNet are implemented. For convenience, TrackNet that takes one input frame is named as Model I and TrackNet that takes three consecutive input frames is named as Model II. For Model II, three consecutive frames are used to detect the ball coordinate in the last frame. During the training phase, three consecutive frames are considered a training sequence if the last frame belongs to the training set. Likewise, three consecutive frames are considered a test sequence if the last frame belongs to the test set. Note that TrackNet framework is scalable. Any number of consecutive input frames are allowed.

\begin{figure}[pth]
\centering
\includegraphics[angle=0, width=3in,keepaspectratio,clip]{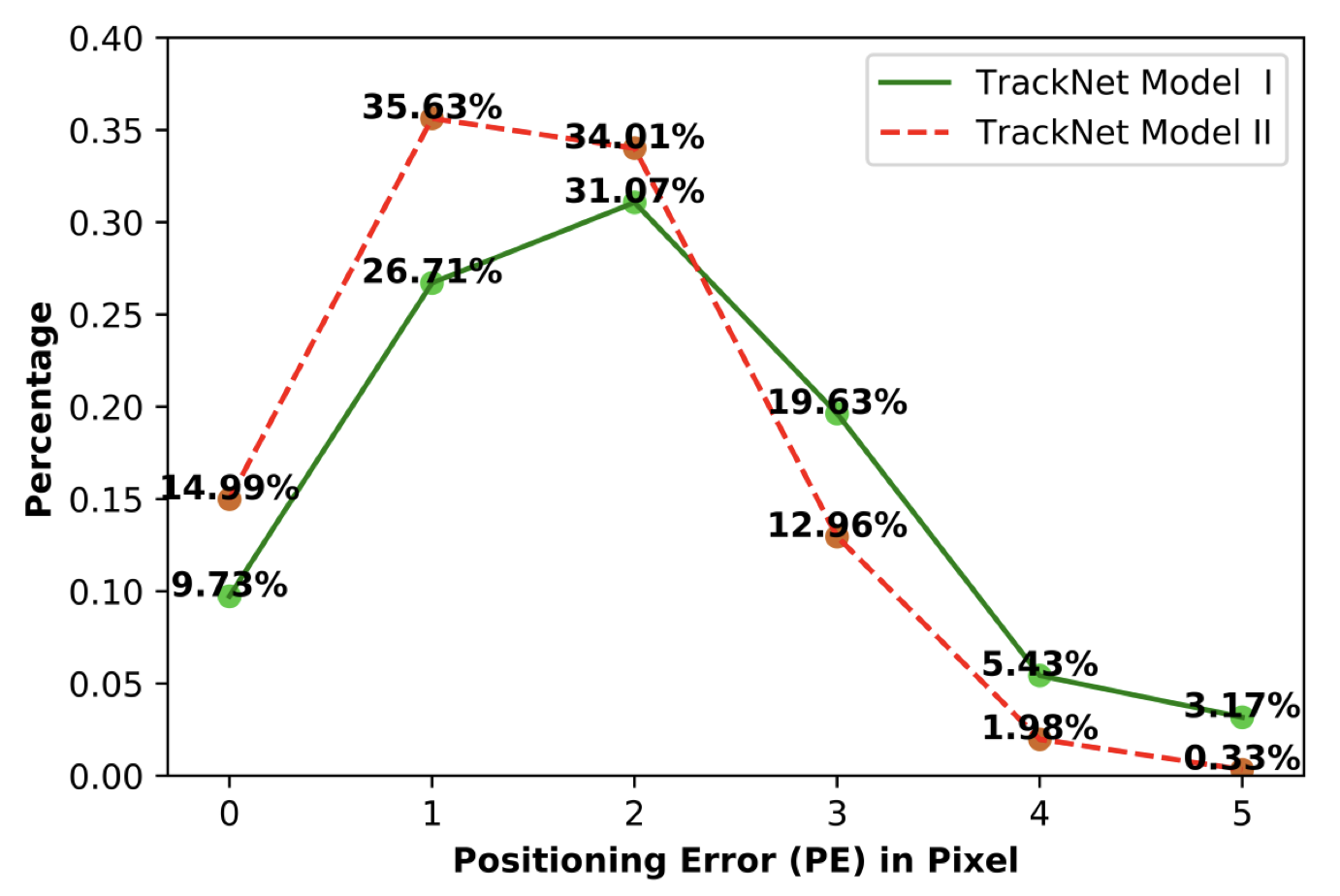}
\hfill\caption{The distribution of the positioning error.}
\label{f_PositioningError}
\end{figure}

To define a proper specification for prediction error, the size of the tennis is investigated. The diameter of tennis images in the video ranges from $2$ to $12$ pixels and the mean diameter is around $5$ pixels. Since the prediction error within a unit size of the ball does not cause misleading in trajectory identification, we define the positioning error (PE) specification as $5$ pixels to indicate whether a ball is accurately detected. Detections with PE larger than $5$ pixels belong to false predictions. $PE$ is defined by the Euclidean distance between the model prediction and the ground truth. Figure \ref{f_PositioningError} shows the $PE$ distribution of TrackNet models. The $x$-axis represents $PE$ in the unit of pixels and the $y$-axis is the percentage of occurrence. $x=0$ stands for perfect detection. $x=1$ means PE lies in $0<PE\leq1$,  $x=2$ means PE lies in $1<PE\leq2$, and so on. Note that occurrence percentages of $PE>5$ of Model I and Model II are $4.3\%$ and $0.1\%$, respectively. That is, $95.7\%$ and $99.9\%$ detections of Model I and Model II fulfill the specification.

The Archana's algorithm \cite{Archana2015Object}, an image processing technique developed by Archana and Geetha, is implemented for comparison. The prediction details of Archana's algorithm, TrackNet Model I, and TrackNet Model II are shown in Table \ref{t_PerformanceSummary}, where TP, FP, TN, and FN stand for true positive, false positive, true negative, and false negative, respectively. The numbers are grouped by "Visibility Class", VC. False positive of VC1, VC2, and VC3 stands for predictions with $PE$ larger than 5 pixels. False negative of VC1, VC2, and VC3 means there is no ball detected or there is more than one ball detected when there is actually one ball in the frame. Note that since TrackNet Model I and TrackNet Model II utilize a different number of input frames, the training set and test set numbers are different. Archanas and TrackNet Model II' follow the same training set and test set as TrackNet Model II.

It is observed that compared to Archana's algorithm, both TrackNet Model I and TrackNet Model II significantly reduce false positives and false negatives, resulting in an increase of both true positives and true negatives. The comparison presents an exceptional object detection capability of deep learning networks over conventional image processing algorithms. In addition, TrackNet Model II performs even better than TrackNet Model I, proving that training TrackNet with consecutive input frames can further improve its dynamic object tracking ability, especially for small objects. Moreover, TrackNet Model II even correctly positions occluded balls occasionally. 2 out of 7 occluded balls are precisely detected. This discovery directly exhibits that consecutive frames provide critical information for the network to learn trajectory patterns of the interested object. By extracting information from neighboring frames, TrackNet Model II not only enhances its tracking precision on normal objects but also on blurry or occluded objects.

The overall performance in terms of precision, recall, and F1-measure are summarized in Table \ref{t_Comparsion}. These three metrics are defined by
\begin{align*}
\text{Precision}  &  =\frac{\text{\# of True Positive}}{\text{\# of True Positive + False Positive}},\\
\text{Recall}  &  =\frac{\text{\# of True Positive}}{\text{\# of VC1+VC2+VC3}}, and\\
\text{F1-measure}  &  =\frac{2(\text{Precision}\times\text{Recall})}{\text{Precision}+\text{Recall}}.
\end{align*}
The Archana's algorithm reaches $92.5\%$ precision, $74.5\%$ recall, and $82.5\%$ F1-measure. With the help of powerful deep learning network, TrackNet Model I outperforms the Archana's algorithm and reaches $95.7\%$ precision, $89.6\%$ recall, and $92.5\%$ F1-measure. By learning how to extract trajectory information from neighboring frames, TrackNet Model II further improves the performance and achieves $99.8\%$ precision, $96.6\%$ recall, and $98.2\%$ F1-measure.

\begin{table}[ptb]
\caption{Accuracy analysis of badminton tracking.}
\label{t_ComparsionBadminton}
\begin{center}
\begin{tabular}
[c]{|l|c|c|c|}\hline
Model & Precision & Recall & F1-measure\\\hline
TrackNet-Tennis& 75.8\% & 22.9\% & 35.2\%\\\hline
TrackNet-Badminton & 85.0\% & 57.7\% & 68.7\%\\\hline
\end{tabular}
\end{center}
\end{table}

To prevent the overfitting issue that frequently happens in deep learning solutions, another $16,118$ frames are added to the training set. These $16,118$ frames are collected from an additional $9$ videos recorded at different tennis courts, including grass court, red clay court, hard court etc. The model trained by the enriched training set is named as TrackNet Model II'. TrackNet Model II' follows the same training logic as TrackNet Model II with the only difference in the variety of training set. The prediction details are shown in Table \ref{t_PerformanceSummary}. As expected, the performance of TrackNet Model II' is similar to TrackNet Model II on the same test set as shown in Table \ref{t_Comparsion}. TrackNet Model II' achieves $99.7\%$ precision, $97.3\%$ recall, and $98.5\%$ F1-measure. Furthermore, 10-fold cross validation is adopted on TrackNet Model II' for the purpose of safety and comprehensive analysis. At last, TrackNet Model II' with 10-fold cross validation reaches $95.3\%$ precision, $75.7\%$ recall, and $84.3\%$ F1-measure.

In addition to tennis, we also apply the proposed TrackNet to the badminton dataset as introduced in Section \ref{sec:Dataset}. The badminton dataset contains $18,242$ frames with the resolution of $1280\times720$. Similarly, all frames are resized from $1280\times720$ to $640\times360$ to speed up the training process. The dataset is randomly divided to the training set and test set. $70\%$ frames are the training set and $30\%$ frames are the test set. For badminton dataset, model training parameters, including learning rate, batch size, number of epochs, etc., are set to the same values used in the training of tennis dataset as shown in Table \ref{t_TrainingParameters}.

Before evaluating TrackNet on badminton, the specification of a correct detection is defined by analyzing the dimension of badminton images in the video. Unlike tennis, badminton is not spherical, resulting in a larger size variation. We define the diameter of a badminton image by taking an average on its largest length and width. The image exists in two extreme cases. One happens when the badminton moves toward the camera at the backcourt and the other happens when the badminton moves laterally at the frontcourt. In our dataset, such cases result in a large variation in images' diameter ranging from $3$ to $24$ pixels. Since the mean diameter is around $7.5$ pixels, we define the PE specification as $7.5$ pixels to indicate whether a badminton image is accurately detected. Detections with PE larger than $7.5$ pixels belong to incorrect predictions. Compared with tennis that has PE specification of $5$ pixels, the PE specification of badminton seems to be released. The main reason is that images of badminton are larger than tennis in the video since the badminton court is smaller than the tennis court. Therefore, the camera uses a smaller focal length to capture the entire court, resulting in larger images of ball and players.

To evaluate the badminton tracking ability of TrackNet, we adopt the transfer learning idea that directly applies the well-trained TrackNet model by tennis dataset for badminton trajectories recognition. Here, we name the transfer learning mode as TrackNet-Tennis which is trained by tennis dataset using three consecutive input frames. As shown in Table \ref{t_ComparsionBadminton}, for badminton tracking, TrackNet-Tennis only achieves precision, recall, and F1-measure of $75.8\%$, $22.9\%$, and $35.2\%$, respectively. Although the precision seems acceptable, the recall is too poor to be used. Such a low recall is due to a large number of false negatives, implying that the badminton cannot be recognized in many circumstances. The main reason causing such poor performance lies in the fundamental characteristics difference between tennis and badminton, including velocity, trajectories, shape, etc. To verify the feasibility of TrackNet framework on badminton tracking, we train another model named as TrackNet-Badminton which is trained by badminton dataset using three consecutive input frames. As shown in Table \ref{t_ComparsionBadminton}, TrackNet-Badminton reaches precision, recall, and F1-measure of $85.0\%$, $57.7\%$, and $68.7\%$, respectively. As expected, TrackNet-Badminton is able to learn the features of badminton, leading to significant performance improvement.

Furthermore, when we compare tennis and badminton tracking performance using TrackNet framework, it can be observed that tennis tracking outperforms badminton tracking by a noticeable margin. This is because badminton travels much faster than tennis, resulting in much more unclear object images in badminton videos. As elaborated in Section \ref{sec:Dataset}, the fastest recorded badminton moves in $417$ kilometers per hour, while the fastest recorded tennis moves in $253$ kilometers per hour. Such an enormous increase in velocity causes performance degradation especially in the aspect of the recall due to high false negatives. High traveling speed makes the badminton move across long distance within only a few frames. The property of dynamic trajectories in such high speed becomes hard to recognize by the model. In addition to the absolute speed, badminton possesses a much higher variation in traveling speed than tennis. For example, in badminton, a drop stroke and a smash stroke have a significant difference in velocity. Such extreme scenarios commonly happen during a badminton competition, making the model hard to fit both scenarios perfectly. Nonetheless, although the performance in tracking badminton is not as phenomenal as tennis, achieving a precision of $85.0\%$ is accurate enough to correctly depict all trajectories in the game. Future research on TrackNet improvement in the aspects of identifying trajectories of extreme fast objects and learning distinct patterns caused by significant speed variation will be conducted.

\section{Conclusion}
\label{sec:Conclusion}
In this paper, we proposed TrackNet, a heatmap-based deep learning network comprising both convolutional and deconvolutional neural network. TrackNet is able to precisely position coordinates of high-speed and tiny objects such as tennis and badminton. With TrackNet, accurate predictions can be achieved on broadcast sports videos without high frame rate and high resolution, significantly reducing the cost from recording and processing high specification videos. To enhance TrackNet's capability of identifying trajectory patterns of fast-moving objects, we designed a scalable input that allows feeding TrackNet with multiple consecutive input frames. By evaluating both conventional image processing algorithm and the proposed TrackNet on the real tennis video dataset, we demonstrated that TrackNet can achieve an explainable and exceptional prediction performance by adopting consecutive input frames concept on the deep neural network. Moreover, for the even faster objects such as badminton, TrackNet achieves a decent tracking capability according to our experimental results, exhibiting promising extensibility to related applications.

\section*{Acknowledgement}
This work of T.-U. \.Ik was supported in part by the Ministry of Science and Technology, Taiwan under grant MOST 107-2627-H-009-001 and MOST 105-2221-E-009-102-MY3.
This work was financially supported by the Center for Open Intelligent Connectivity from The Featured Areas Research Center Program within the framework of the Higher Education Sprout Project by the Ministry of Education (MOE), Taiwan.

\bibliographystyle{IEEEtran}
\balance
\bibliography{Tennis}
\end{document}